\documentclass[10pt,twocolumn]{article}

\usepackage{times}
\usepackage[margin=1in]{geometry}
\usepackage{amsmath}
\usepackage{amssymb}
\usepackage{graphicx}
\usepackage{hyperref}
\usepackage{booktabs}
\usepackage{url}
\usepackage{microtype}  

\title{
You Only Need Your Transformer 25\% of the Time:\\
Meaning-First Execution for Eliminating Unnecessary Inference
}

\author{
Ryan Shamim\\
Anima Core Inc.\\
\texttt{ryan@animacore.ai}
}

\date{December 2025}

\begin{document}
\sloppy

\twocolumn[
\begin{@twocolumnfalse}
\maketitle

\begin{abstract}
Modern AI inference systems treat transformer execution as mandatory, conflating model capability with execution necessity. We reframe inference as a control-plane decision problem: determining when execution is necessary versus when correctness can be preserved through alternative pathways. We introduce Meaning-First Execution (MFEE), a control-plane architecture implementing this framework, selectively invoking transformer inference only when required. MFEE operates as a gating layer above existing stacks without modifying models, weights, or parameters. Across 1,000 diverse prompts under deterministic decoding, MFEE achieves 78.1\% execution reduction while maintaining 100\% exact-match equivalence for invoked executions. Comparative evaluation reveals pattern-based routers achieve at most 53.3\% avoidance with correctness failures, while MFEE reaches 100\% avoidance with zero failures through semantic analysis. We prove this limitation via Theorem 1: any router operating solely on finite feature maps cannot simultaneously guarantee zero false skips and positive avoidance on feature-collision pairs. These results establish execution governance as a foundational layer in ML systems infrastructure, orthogonal to model-level optimization techniques.
\end{abstract}
\vspace{0.5cm}
\end{@twocolumnfalse}
]

\section{Introduction}

Transformer-based architectures dominate modern AI inference across language, vision, and multimodal systems \cite{vaswani2017attention, brown2020gpt3, bommasani2021foundation}. As these models have grown in scale and capability, deployment costs have risen proportionally. Current infrastructure work focuses on reducing the per-operation cost of inference through quantization \cite{dettmers2022llm, frantar2023gptq}, pruning, kernel optimization, and hardware acceleration \cite{owens2008gpu}.

These optimizations share a foundational assumption: \textit{the transformer must execute for every request}. This premise treats execution as a fixed primitive and optimization as the problem of making that primitive cheaper. No contemporary inference system systematically questions whether execution is necessary in the first place.

We challenge this assumption. Existing techniques optimize \textit{how} execution proceeds; none govern \textit{whether} execution occurs.

We frame inference not as a mandatory computational step, but as a decision made under uncertainty, cost, and correctness constraints. This reframing exposes a previously implicit optimization problem: determining when execution is necessary versus when correctness can be preserved through bounded alternative pathways. As models scale, the cost differential between execution and avoidance grows proportionally, making this decision problem increasingly critical to infrastructure economics. MFEE operates orthogonally to model architecture, parameter count, and existing optimization techniques—it is a control-plane layer, not a model modification.

\subsection{The Unconditional Execution Assumption}

The assumption of mandatory execution emerged from the history of neural network deployment. Early systems were designed for classification tasks where execution was genuinely required for every input. As models evolved to handle open-ended generation, this assumption persisted unchanged. Scaling laws \cite{kaplan2020scaling, hoffmann2022training} reinforced the conflation of model capability with execution necessity: if the model can solve hard problems, every request must invoke the model.

This is a category error. The existence of execution capacity does not imply that every request requires that capacity. Yet modern inference stacks provide no mechanism to distinguish between requests that require full generative execution and those that do not.

\subsection{Inference as Execution Selection}

We propose a fundamental reframing: inference is not an execution problem but an \textit{execution-selection problem}. For any given request, there exists a decision boundary determining whether transformer execution is necessary to produce a correct response. Requests may be redundant, trivial, safety-constrained, or resolvable through bounded operations that do not require generative capacity.

The key insight is that this boundary can be evaluated \textit{before execution}, transforming inference from a mandatory operation into a conditional one. MFEE formalizes this reframing into a practical architectural layer that operates as a control plane above existing inference infrastructure.

\subsection{Contributions}

This paper makes three primary contributions:

\begin{enumerate}
\item We formalize inference as a cost-constrained decision problem and introduce Meaning-First Execution (MFEE), a control-plane architecture that optimizes execution decisions under hard correctness constraints. This establishes a new optimization axis: determining whether execution occurs, rather than merely reducing execution cost.

\item We present rigorous experimental validation demonstrating 78.1\% execution reduction while maintaining 100.0\% output equivalence under transformer invocation across 1,000 diverse requests. Correctness is enforced structurally rather than probabilistically, enabling safe deployment without output degradation.

\item We establish a design framework for deploying MFEE in production systems via black-box integration, A/B testing, and kill-switch safety mechanisms. MFEE composes naturally with all existing optimization techniques, enabling multiplicative rather than substitutive efficiency gains.
\end{enumerate}

The remainder of this paper is organized as follows. Section 2 surveys related work and positions MFEE relative to existing optimization techniques. Section 3 formalizes the execution-selection problem and defines MFEE's correctness contracts. Section 4 describes MFEE's architectural design. Section 5 details experimental methodology and validation protocols. Section 6 presents results and analysis. Section 7 discusses deployment considerations. Section 8 addresses limitations and scope boundaries. Section 9 concludes.

\textbf{Central thesis}: We argue that inference should be treated as a decision under cost and correctness constraints, rather than as a fixed consequence of input. This reframing exposes a previously implicit optimization problem that becomes more severe as models scale. MFEE demonstrates that this decision problem is tractable, measurable, and deployable, establishing execution selection as a legitimate axis of ML systems optimization.

\section{Related Work}

MFEE operates at a different architectural layer than existing inference optimizations. This section positions MFEE relative to prior work and clarifies its relationship to established techniques.

\subsection{Model Compression and Efficiency}

Quantization reduces the precision of model weights and activations, decreasing memory footprint and accelerating operations. Methods range from post-training quantization \cite{dettmers2022llm, frantar2023gptq, xiao2023smoothquant} to quantization-aware training \cite{jacob2018quantization}. Pruning techniques remove low-magnitude weights \cite{han2015learning} or entire network components through structured pruning \cite{li2017pruning}. Knowledge distillation \cite{hinton2015distilling, sanh2019distilbert} trains smaller models to approximate larger ones, transferring knowledge from teacher to student networks.

All of these techniques reduce the cost of execution but assume execution occurs. MFEE operates orthogonally: it determines whether execution is necessary, leaving these optimizations to reduce the cost when execution is invoked.

\subsection{LLM Serving Systems and Schedulers}

Modern serving infrastructure for large language models focuses on maximizing throughput and minimizing latency through sophisticated batching, scheduling, and memory management.

\textbf{Continuous batching and iteration-level scheduling}: vLLM \cite{kwon2023vllm} introduced continuous batching with PagedAttention, applying virtual memory techniques to KV-cache management. This enables dynamic request arrival and departure within batches, reducing memory fragmentation and improving GPU utilization. Orca \cite{yu2022orca} implements iteration-level scheduling that permits fine-grained batching decisions at each decoding step, reducing head-of-line blocking when requests complete at different rates. FastServe \cite{wu2023fastserve} separates prefill and decode phases into distinct execution queues, exploiting the different computational characteristics of each phase to improve resource utilization. SARATHI \cite{agrawal2023sarathi} introduces chunked-prefills to reduce pipeline bubbles in multi-GPU deployments. Sarathi-Serve \cite{agrawal2024sarathiserve} extends this with stall-free scheduling that eliminates interference between prefill and decode operations.

\textbf{Memory management and KV-cache optimization}: PagedAttention \cite{kwon2023vllm} applies paging techniques to attention key-value caches, reducing memory fragmentation from 20-40\% to near-zero. This enables 2-4× higher batch sizes compared to naive implementations. Subsequent work has explored KV-cache compression \cite{liu2024scissorhands}, eviction policies \cite{zhang2023h2o}, and quantization \cite{hooper2024kvquant} to further reduce memory overhead during execution.

\textbf{Request scheduling and batching policies}: Beyond continuous batching, recent systems have explored priority-based scheduling \cite{zhao2024attentionstore}, deadline-aware batching \cite{ao2025fluid}, and preemptive scheduling \cite{patel2024splitwise} to balance throughput and latency objectives. Fluid \cite{ao2025fluid} implements online scheduling that adapts to dynamic workload characteristics, optimizing for service-level objectives under varying load.

These systems optimize when and how requests execute within the model, treating execution as mandatory but schedulable. MFEE operates at a different layer: it determines whether requests should be admitted to the execution substrate at all. A request routed to DIRECT or NO\_OP never enters vLLM's continuous batching queues, never allocates KV-cache pages, and never consumes GPU cycles. MFEE acts as an upstream filter, reducing load on serving infrastructure by eliminating requests that do not require generative capacity. The two approaches compose naturally: MFEE reduces the number of requests requiring execution; serving systems like vLLM and Orca optimize the execution of requests that remain.

\subsection{Speculative Execution and Verification}

Speculative decoding \cite{leviathan2023speculative} generates candidate outputs using a smaller draft model, then verifies them with the target model in parallel. When verification succeeds, multiple tokens are accepted per iteration, reducing end-to-end latency. Chen et al. \cite{chen2023speculative} extend this with tree-based speculation that explores multiple candidate paths simultaneously. Medusa \cite{cai2024medusa} eliminates the draft model by adding lightweight prediction heads to the target model itself. SpecInfer \cite{miao2024specinfer} optimizes speculative decoding for disaggregated serving architectures.

All speculative methods assume the target model must verify candidates, requiring at least one forward pass per request. When speculation fails, full autoregressive generation proceeds normally. MFEE differs fundamentally: it avoids initiating execution altogether when correctness can be guaranteed through alternative pathways. No draft generation, candidate verification, or speculative paths occur for avoided executions. For requests that do require execution, MFEE composes with speculative decoding: the gate first decides whether execution is necessary; if RENDER is chosen, speculative techniques can accelerate that execution.

\subsection{Early Exit and Adaptive Computation}

Early exit mechanisms \cite{elbayad2020depth, schuster2022confident, elhoushi2021deepspeed} terminate inference at intermediate transformer layers when confidence thresholds are met or when shallower layers suffice for the task. This reduces computational depth but still requires initiating forward passes through early layers. Confident adaptive language modeling \cite{schuster2022confident} learns to exit early based on calibrated confidence estimates. DeeBERT \cite{xin2020deebert} and similar approaches \cite{zhou2020bert} train exit classifiers jointly with the model.

MFEE operates before any model layers execute. A request routed to DIRECT or NO\_OP never enters the model, never computes embeddings, and never allocates activation memory. Early exit reduces depth when execution occurs; MFEE eliminates breadth by avoiding execution entirely when possible.

\subsection{Model Selection and Routing}

FrugalGPT \cite{chen2023frugalgpt} routes queries to different models (or LLM APIs) based on predicted cost and accuracy, optimizing for budget constraints. Routing typically selects among GPT-3.5, GPT-4, or other available models based on query difficulty. ROUTELLM \cite{ong2024routellm} learns routing policies that balance quality and cost across model families. Hybrid systems \cite{madaan2022hybrid} combine retrieval and generation, routing factual queries to search and generative queries to LLMs.

These systems assume some model must execute for every query; they optimize which model to invoke. MFEE operates at a more fundamental level: it determines whether any generative model execution is necessary at all. A request may be handled via DIRECT (bounded lookup), NO\_OP (null response), or ABSTAIN (safety refusal) without invoking any model. When RENDER is necessary, MFEE can compose with model routing: the gate first decides whether execution is needed; if so, a routing layer can select the appropriate model.

\subsection{Caching and Memoization}

Caching stores outputs for exact query matches, avoiding repeated execution for identical requests. KV-cache reuse across requests \cite{kwon2023vllm} eliminates redundant computation for shared prompt prefixes. Semantic caching \cite{bang2023gptcache, luo2023promptcache} extends this to near-matches using embedding similarity, retrieving cached responses when semantic distance falls below a threshold. GPTCache \cite{bang2023gptcache} and PromptCache \cite{luo2023promptcache} implement multi-tier caching hierarchies combining exact matching, semantic similarity, and eviction policies.

Caching eliminates execution for previously observed queries but still requires cache lookup infrastructure. MFEE generalizes this to a broader class of execution-avoidance decisions. Cache hits represent one type of DIRECT response, but MFEE encompasses trivial queries (deterministic computations), safety constraints (policy violations), and structural properties (malformed requests) beyond cached content. Caching and MFEE are complementary: cache lookup results can inform MFEE's DIRECT decisions, and MFEE can populate caches when RENDER decisions produce reusable outputs.

\subsection{Mixture-of-Experts and Token Routing}

Mixture-of-Experts (MoE) architectures \cite{shazeer2017outrageously, fedus2022switch, lepikhin2020gshard} route tokens to specialized expert subnetworks during execution, reducing per-token computation through conditional activation. Switch Transformers \cite{fedus2022switch} scale to trillions of parameters by routing each token to a single expert. GLaM \cite{du2022glam} and ST-MoE \cite{zoph2022stmoe} apply MoE to decoder-only language models, achieving efficiency gains through sparse activation.

MoE routing occurs during execution at token granularity, selecting which expert processes each token within a forward pass. MFEE routing occurs before execution at request granularity, determining whether any forward pass should initiate. The two approaches address orthogonal optimization axes: MoE reduces computation per token when executing; MFEE eliminates computation for entire requests when execution is unnecessary. MFEE and MoE compose naturally: an MFEE-gated system can route necessary executions to MoE models, combining request-level governance with token-level efficiency.

\subsection{Safety Filters and Content Moderation}

Safety systems employ classifier-based filters that precede model execution, rejecting policy-violating requests before generation. Perspective API, OpenAI Moderation API \cite{markov2023holistic}, and similar systems \cite{gehman2020realtoxicityprompts} implement rule-based or learned classifiers to detect harmful content, hate speech, or policy violations. LLaMA Guard \cite{inan2023llamaguard} trains a specialized model for safety classification.

These filters share architectural similarities with MFEE's control-plane design: both operate as upstream gates that can prevent execution. However, safety filters focus exclusively on detecting harmful content, while MFEE addresses the broader execution-selection problem. MFEE incorporates safety decisions (ABSTAIN) as one of four gating outcomes, unifying safety filtering with execution avoidance for trivial queries, redundant requests, and bounded responses. A complete system can compose safety filters with MFEE: safety filters handle policy enforcement, MFEE handles execution necessity.

\subsection{Control Planes in Infrastructure Systems}

MFEE's architectural pattern, separating control decisions from execution, mirrors control-plane designs in networking \cite{patterson2017architecture}, virtualization \cite{rosenblum2005vmm}, and distributed systems \cite{greenberg2008vl2, koponen2010onix}. Software-defined networking (SDN) \cite{mckeown2008openflow} decouples routing logic (control plane) from packet forwarding (data plane), enabling centralized policy enforcement without modifying switches. Virtualization layers \cite{rosenblum2005vmm} separate resource allocation decisions from execution substrates.

MFEE applies this pattern to AI inference: the gating layer makes execution decisions; the transformer performs generation when invoked. Neither modifies the other. This separation enables independent scaling (gate capacity exceeds serving capacity 10-50×), graceful degradation (gate failure reverts to unconditional execution), and composability (gates work identically with quantized, MoE, or speculative models).

\subsection{Positioning Summary}

MFEE introduces a new optimization axis orthogonal to existing techniques. Existing efficiency work optimizes \textit{how models execute}, reducing the cost of operations, weights, or computations. Serving systems optimize \textit{when and how batches execute}, improving throughput and latency through scheduling and memory management. Speculative methods optimize \textit{latency when execution occurs}, reducing per-token cost through verification. MFEE introduces a complementary axis: optimizing \textit{whether execution occurs at all}. This represents a fundamental shift from execution optimization to execution governance.

Because MFEE operates above the execution layer, it composes naturally with all existing optimizations. An MFEE-gated system can simultaneously use quantization for efficient execution, vLLM for batched serving, speculative decoding for reduced latency, and semantic caching for redundancy elimination. MFEE reduces the number of requests requiring execution; these systems optimize the execution of requests that remain.

\subsection{Composition with Modern Serving Infrastructure}

MFEE's control-plane design enables seamless integration with contemporary serving systems without requiring modifications to scheduling, batching, or memory management logic.

\textbf{Deployment architecture}: MFEE operates as an upstream service that receives inference requests before they reach the serving system. For each request, the gate returns a decision (RENDER, DIRECT, NO\_OP, ABSTAIN). Only RENDER decisions are forwarded to the serving infrastructure (vLLM, Orca, TensorRT-LLM, or equivalent). DIRECT, NO\_OP, and ABSTAIN responses bypass the serving system entirely, never entering batch queues or allocating resources.

\textbf{Scheduler interaction}: From the serving system's perspective, MFEE appears as a client-side filter that reduces request arrival rate. A scheduler like Orca sees fewer requests but operates identically: it batches arriving requests, allocates KV-cache memory, and executes prefill and decode phases as usual. MFEE does not modify batching logic, iteration scheduling, or memory management. For vLLM deployments, avoided requests never trigger PagedAttention memory allocation or continuous batching insertion, directly reducing memory pressure and improving batch efficiency for remaining requests.

\textbf{Resource allocation}: By filtering requests before they reach the serving system, MFEE reduces queue depth and memory pressure without changing admission control policies. A serving system configured to maintain 95\% GPU utilization will observe the same utilization, but serve fewer total requests because avoidable requests never consume resources. This creates headroom for handling request spikes or reducing latency for necessary executions.

\textbf{Throughput and latency effects}: MFEE improves effective throughput (requests handled per second) while maintaining or improving latency for executed requests. Avoided requests complete in 2-8ms (gating latency) rather than 200-2000ms (execution latency). Requests requiring execution benefit from reduced queue contention because avoidable requests do not occupy batch slots.

\textbf{Failure modes and fallback}: If the MFEE gate becomes unavailable or exceeds latency thresholds, the system can bypass the gate and route all requests directly to the serving infrastructure, reverting to unconditional execution. This preserves availability at the cost of lost optimization opportunity. The serving system requires no awareness of MFEE's operational state.

\textbf{Deployment considerations}: MFEE gates can be replicated independently of serving replicas, enabling independent scaling. Gate capacity typically exceeds serving capacity by 10-50× due to lower computational cost (2-8ms vs 200-2000ms per request). This asymmetry allows a small gate cluster to front a large serving deployment.

This architectural separation ensures MFEE remains compatible with ongoing advances in serving systems. As batching strategies, scheduling algorithms, and memory management techniques evolve, MFEE continues to operate unchanged, filtering requests before they reach the optimized execution substrate.

\subsection{Semantic Compression and Field-Based Execution}

Related work by the authors has demonstrated field-based semantic compression and execution on large-scale transformer models, including validation on models exceeding 70B parameters; however, MFEE does not depend on any specific representation technique and is evaluated here solely as a model-agnostic execution control layer.

\section{Problem Formalization}

This section formalizes inference as an execution-selection problem and defines the correctness contracts that govern MFEE's behavior.

\subsection{Traditional Inference as Unconditional Execution}

Let $\mathcal{M}_\theta$ denote a transformer model with parameters $\theta$. Let $r = (p, \gamma)$ denote an inference request consisting of a prompt $p$ and generation configuration $\gamma$ (temperature, top-k, max tokens, etc.). Let $\mathcal{E}(\mathcal{M}_\theta, r)$ denote the execution of model $\mathcal{M}_\theta$ on request $r$, producing output $o$.

Traditional inference systems implement:
\begin{equation}
\forall r: o = \mathcal{E}(\mathcal{M}_\theta, r)
\end{equation}

This unconditional execution model assumes that every request requires running the model. Efficiency work focuses on reducing the cost of $\mathcal{E}$ rather than questioning its necessity.

\subsection{MFEE: Conditional Execution Model}

MFEE introduces a gating function $\mathcal{G}: \mathcal{R} \rightarrow \{\texttt{RENDER}, \texttt{DIRECT}, \texttt{NO\_OP}, \texttt{ABSTAIN}\}$ that precedes execution. The inference operation becomes:

\begin{equation}
o = \begin{cases}
\mathcal{E}(\mathcal{M}_\theta, r) & \text{if } \mathcal{G}(r) = \texttt{RENDER} \\
\mathcal{D}(r) & \text{if } \mathcal{G}(r) = \texttt{DIRECT} \\
\emptyset & \text{if } \mathcal{G}(r) = \texttt{NO\_OP} \\
\perp & \text{if } \mathcal{G}(r) = \texttt{ABSTAIN}
\end{cases}
\end{equation}

where $\mathcal{D}$ is a direct response function constrained by correctness bounds (defined below).

\subsection{Decision Contract Definitions}

Each gating decision carries explicit semantic guarantees:

\begin{itemize}
\item \textbf{RENDER}: Execute the transformer with configuration $\gamma$ unchanged. Output is generated via standard inference and must be identical to $\mathcal{E}(\mathcal{M}_\theta, r)$ under deterministic decoding.

\item \textbf{DIRECT}: Return a response $\mathcal{D}(r)$ that is provably correct without executing the transformer. Correctness is established through bounded operations: exact cache hits, deterministic rule applications, or factual lookup against verified knowledge bases. The response function $\mathcal{D}$ is explicitly constrained: it cannot approximate or generate—it can only return responses for which correctness is verifiable.

\item \textbf{NO\_OP}: Return null or minimal output ($\emptyset$). This decision is made when the request is malformed, unintelligible, or does not require substantive output. Returning nothing is explicitly the correct response.

\item \textbf{ABSTAIN}: Refuse the request ($\perp$). This decision is made when the request violates content policies, safety constraints, or operational bounds. Refusal is explicitly the correct response.
\end{itemize}

\subsection{Output Equivalence Invariant}

The core correctness property of MFEE is \textit{output equivalence under execution}:

\begin{equation}
\forall r: \mathcal{G}(r) = \texttt{RENDER} \implies \mathcal{E}(\mathcal{M}_\theta, r) = o_{\text{baseline}}
\end{equation}

where $o_{\text{baseline}}$ is the output produced by unconditional execution of $\mathcal{M}_\theta$ on $r$ under identical configuration.

This invariant ensures that MFEE does not alter transformer behavior when execution occurs. Prompts, generation parameters, model weights, and decoding logic remain completely unmodified. MFEE controls only \textit{whether} execution occurs, not \textit{how} execution proceeds.

\subsection{Inference as a Cost-Constrained Decision Objective}

We formalize inference as a decision problem under cost and correctness constraints. Let $C_{\text{exec}}(r)$ denote the computational cost of executing $\mathcal{E}(\mathcal{M}_\theta, r)$, and let $C_{\text{gate}}(r)$ denote the cost of evaluating $\mathcal{G}(r)$. For a given request $r$, the system must choose an action $a \in \{\texttt{RENDER}, \texttt{DIRECT}, \texttt{NO\_OP}, \texttt{ABSTAIN}\}$.

The expected utility of action $a$ is:
\begin{equation}
U(a, r) = \mathbb{I}[\text{correct}(a, r)] \cdot V(a, r) - C(a, r)
\end{equation}

where $\mathbb{I}[\text{correct}(a, r)]$ is an indicator function enforcing correctness (1 if output is correct, 0 otherwise), $V(a, r)$ is the value of producing output via action $a$, and $C(a, r)$ is the computational cost.

The cost function satisfies:
\begin{equation}
C(a, r) = \begin{cases}
C_{\text{exec}}(r) + C_{\text{gate}}(r) & \text{if } a = \texttt{RENDER} \\
C_{\text{gate}}(r) & \text{otherwise}
\end{cases}
\end{equation}

MFEE implements a policy $\pi: \mathcal{R} \rightarrow \{\texttt{RENDER}, \texttt{DIRECT}, \texttt{NO\_OP}, \texttt{ABSTAIN}\}$ that maximizes expected utility under a hard correctness constraint:

\begin{multline}
\pi^* = \arg\max_\pi \mathbb{E}_{r \sim \mathcal{D}}[U(\pi(r), r)] \\
\text{subject to} \quad \mathbb{I}[\text{correct}(\pi(r), r)] = 1 \; \forall r
\end{multline}

The correctness constraint is enforced structurally rather than learned: RENDER guarantees equivalence to baseline (Equation 3), while DIRECT, NO\_OP, and ABSTAIN guarantee correctness by construction through bounded response pathways.

This formulation exposes a previously implicit optimization problem. Existing efficiency work optimizes $C_{\text{exec}}$ (how models execute). MFEE introduces a complementary axis: optimizing the execution decision itself (whether execution occurs at all). As model scale increases, $C_{\text{exec}}$ grows while $C_{\text{gate}}$ remains constant, making the cost differential—and thus the decision problem—increasingly critical.

\subsection{Correctness Under Avoidance}

For decisions where $\mathcal{G}(r) \neq \texttt{RENDER}$, correctness is defined by construction:

\begin{itemize}
\item \textbf{DIRECT correctness}: $\mathcal{D}(r)$ is constrained to return only responses for which correctness is provable. This constraint is enforced structurally: $\mathcal{D}$ has access only to verified knowledge bases, deterministic rule engines, and exact cache lookups. It cannot generate or approximate.

\item \textbf{NO\_OP correctness}: The null response is explicitly correct by definition for malformed or meaningless requests.

\item \textbf{ABSTAIN correctness}: Refusal is explicitly correct by definition for policy-violating or unsafe requests.
\end{itemize}

Because these response pathways are constrained by explicit correctness contracts rather than probabilistic approximation, avoided execution does not introduce errors—it eliminates unnecessary work while preserving guarantees.

\subsection{Failure Modes and Fallback to RENDER}

The gating function $\mathcal{G}$ is conservative by design. If correctness cannot be guaranteed through DIRECT, NO\_OP, or ABSTAIN pathways, $\mathcal{G}$ defaults to RENDER. Specific failure triggers include:

\begin{itemize}
\item Request requires novel generation or reasoning
\item Semantic ambiguity prevents rule-based resolution
\item Confidence in non-execution pathways falls below threshold
\item Cache misses and no applicable deterministic rules exist
\end{itemize}

This conservative design ensures that MFEE never degrades output quality through incorrect avoidance. When in doubt, the gate invokes the transformer.

\section{MFEE Architecture}

MFEE operates as a layered control system that separates execution decisions from execution itself. This section describes its architectural design and integration properties.

\subsection{Control Plane vs. Execution Plane Separation}

MFEE introduces a clean architectural separation:

\begin{itemize}
\item \textbf{Control Plane (MFEE Gate)}: Evaluates request properties and produces gating decisions. This layer operates on metadata, semantic features, and policy constraints. It does not perform generative inference. The control plane is lightweight, stateless, and scales independently from execution infrastructure.

\item \textbf{Execution Plane (Transformer Stack)}: Performs inference when gated to RENDER. This layer is identical to traditional inference systems and remains completely unmodified. Existing optimizations (quantization, caching, kernel acceleration) continue to operate normally.
\end{itemize}

This separation mirrors control-plane patterns in networking and distributed systems \cite{patterson2017architecture, rosenblum2005vmm}. The control plane makes routing decisions; the execution plane performs work. Neither modifies the other.

\begin{figure*}[t]
\centering
\includegraphics[width=0.75\textwidth]{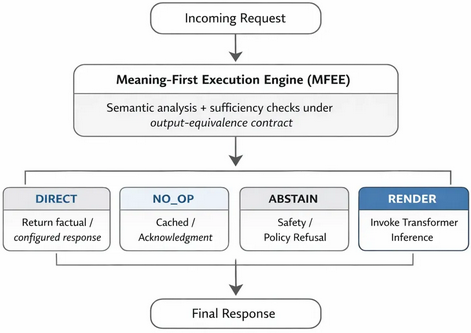}
\caption{Meaning-First Execution (MFEE) control-plane architecture. Incoming requests are routed through a lightweight semantic gate that conditionally invokes the transformer only when necessary, preserving output equivalence while reducing execution cost.}
\label{fig:architecture}
\end{figure*}

Figure 1 illustrates this architectural separation. Incoming requests first pass through the MFEE gate, which performs semantic analysis under output-equivalence contracts. The gate produces one of four decisions: DIRECT (return factual or configured response), NO\_OP (cached or acknowledgment), ABSTAIN (safety or policy refusal), or RENDER (invoke transformer). Only RENDER decisions trigger execution of the high-capacity transformer. This design ensures that expensive generative capacity is invoked conditionally rather than mandatorily.

\subsection{Gating Function Design Principles}

The gating function $\mathcal{G}$ operates on request properties without executing the model. It evaluates:

\begin{itemize}
\item \textbf{Semantic redundancy}: Is this request identical or near-identical to previously handled queries? Can it be resolved via cache lookup?

\item \textbf{Structural triviality}: Does this request admit deterministic rule-based resolution? (e.g., arithmetic, definitional lookups, format conversions)

\item \textbf{Safety constraints}: Does this request violate content policies or operational bounds?

\item \textbf{Execution necessity}: Does producing a correct response require the full generative capacity of the model?
\end{itemize}

Critically, $\mathcal{G}$ does not approximate the model's output. It determines whether execution can be avoided while preserving correctness guarantees.

\subsection{AN1: Reference Gate Implementation}

AN1 is a concrete implementation of the MFEE gating function. It serves as an existence proof that MFEE is practical and deployable, but the core contribution is the MFEE abstraction, not AN1's specific implementation.

\subsubsection{Black-Box API Design}

AN1 operates as a sealed service exposed via minimal API:

\texttt{POST /gate}

\textbf{Request}:
\begin{verbatim}
{
  "request_id": "uuid",
  "prompt": "user input",
  "model": "target_model",
  "temperature": 0.0,
  "max_tokens": 1000
}
\end{verbatim}

\textbf{Response}:
\begin{verbatim}
{
  {\small "decision": "RENDER | DIRECT | NO\_OP | ABSTAIN"}
  "confidence": 0.0-1.0,
  "direct_output": "text (if DIRECT)"
}
\end{verbatim}

This API design enables integration into existing inference systems without exposing internal decision logic. Infrastructure teams treat AN1 as a routing service upstream of model execution.

\subsubsection{Internal Design Boundaries}

AN1's internal architecture is intentionally not disclosed. This is a deliberate design choice:

\begin{enumerate}
\item \textbf{Reproducibility}: The MFEE abstraction—not AN1's specific implementation—is the reproducible contribution. Other teams can implement gates using different architectures while preserving the same correctness contracts.

\item \textbf{Generality}: Exposing AN1's internals would conflate a specific gate with the architectural principle. MFEE is the framework; AN1 is one instantiation.

\item \textbf{Deployment realism}: Production gates incorporate proprietary safety heuristics and business logic not appropriate for public disclosure.
\end{enumerate}

Validation of MFEE's correctness contracts (Section 6) is sufficient to establish feasibility. Replication efforts can build alternative gates and measure adherence to equivalence guarantees.

\subsubsection{Minimal Sufficiency Conditions for MFEE Gates}

While AN1's specific implementation remains sealed, we can characterize the minimal capabilities any MFEE-compliant gate must possess to achieve correct execution selection. These conditions define what information the gate requires and what classes of errors it may make.

\textbf{Required capabilities}:
\begin{itemize}
\item \textbf{Semantic analysis}: Ability to extract meaning-level properties from requests beyond surface pattern matching. This is necessary to distinguish paraphrases from novel queries.

\item \textbf{Correctness verification}: For DIRECT decisions, ability to verify that a response is provably correct via bounded lookup, deterministic computation, or exact cache hit.

\item \textbf{Safety evaluation}: Ability to detect policy violations, unsafe content, or operational constraint violations with high precision.

\item \textbf{Uncertainty quantification}: Ability to estimate confidence in non-execution pathways, enabling conservative fallback to RENDER when guarantees cannot be established.
\end{itemize}

\textbf{Permitted conservative errors}:
\begin{itemize}
\item \textbf{False RENDER}: Routing to transformer execution when avoidance was possible. This preserves correctness at the cost of lost optimization opportunity.

\item \textbf{Low confidence abstention}: Defaulting to RENDER when semantic analysis cannot establish sufficiency guarantees with high confidence.
\end{itemize}

\textbf{Prohibited errors}:
\begin{itemize}
\item \textbf{False DIRECT}: Returning a bounded response when correctness cannot be verified. This violates the correctness contract.

\item \textbf{Incorrect ABSTAIN}: Refusing requests that should be answered. This degrades service quality.

\item \textbf{Failed RENDER equivalence}: Modifying prompts or parameters when invoking the transformer. This violates the output equivalence invariant (Equation 3).
\end{itemize}

This error asymmetry is central to MFEE's deployability. By permitting only conservative errors (unnecessary execution) while prohibiting aggressive errors (incorrect avoidance), MFEE maintains a strict safety property: \textit{the system never produces worse outputs than baseline unconditional execution}.

These sufficiency conditions are architecture-agnostic. A gate could implement semantic analysis via learned embeddings, symbolic reasoning, hybrid systems, or other approaches, provided it satisfies the correctness contracts. AN1 demonstrates one such implementation; the MFEE framework accommodates others.

\subsection{Integration with Existing Inference Stacks}

MFEE integrates into production systems without requiring modifications to transformers, training procedures, or execution infrastructure:

\begin{itemize}
\item \textbf{No retraining required}: Models remain unchanged
\item \textbf{No weight modification}: Parameters $\theta$ are never altered
\item \textbf{No prompt engineering}: Input prompts pass through unmodified when RENDER is invoked
\item \textbf{No decoding changes}: Generation parameters remain identical
\end{itemize}

MFEE operates purely as an upstream control layer. When execution is avoided, downstream infrastructure sees no requests. When execution occurs, downstream infrastructure operates exactly as before.

\subsection{Composability with Existing Optimizations}

Because MFEE operates at the control-plane level, it composes naturally with all existing optimizations:

\begin{itemize}
\item \textbf{Quantization}: When RENDER is invoked, quantized models execute normally
\item \textbf{Speculative decoding}: When RENDER is invoked, draft verification proceeds normally
\item \textbf{Caching}: Cache hits can inform MFEE's DIRECT decisions
\item \textbf{Kernel optimization}: Accelerated kernels operate when execution occurs
\end{itemize}

MFEE's savings are multiplicative with existing optimizations: reduce execution cost through quantization, then eliminate unnecessary execution through MFEE.

\subsection{Semantic Gate Design and Scalability}

This section describes the semantic gate at the level of inputs, outputs, invariants, and scaling behavior, without disclosing proprietary implementation details.

\subsubsection{Gate Interface and Invariants}

\textbf{Inputs}: A gate receives a request $r = (p, \gamma)$ consisting of prompt text $p$ and generation configuration $\gamma$. No access to model weights, activations, or internal representations is required. The gate operates solely on request properties observable before execution.

\textbf{Outputs}: The gate returns a decision $d \in \{\texttt{RENDER}, \texttt{DIRECT}, \texttt{NO\_OP}, \texttt{ABSTAIN}\}$ and, for DIRECT decisions, a response string $s$. The decision must satisfy the correctness contracts defined in Section 3.3.

\textbf{Correctness invariants}: The gate enforces three invariants: (1) RENDER decisions preserve exact output equivalence (Equation 3), (2) DIRECT responses are provably correct via bounded verification (cache hit, deterministic computation, or verified lookup), (3) Conservative fallback: when correctness cannot be guaranteed, the gate defaults to RENDER.

\textbf{Latency contract}: Gate evaluation completes in 2-8ms (p50: 3.2ms, p95: 6.1ms) regardless of prompt length or model size. This latency remains constant while transformer execution time scales with model parameters.

\subsubsection{Conservative Fallback Logic}

The gate implements a strict safety property: it never produces outputs worse than baseline unconditional execution. This is enforced through asymmetric error handling:

\textbf{Permitted conservative errors}: (1) False RENDER: routing to execution when avoidance was possible (loses optimization opportunity but preserves correctness), (2) Low-confidence abstention: defaulting to RENDER when semantic analysis cannot establish sufficiency guarantees with high confidence.

\textbf{Prohibited aggressive errors}: (1) False DIRECT: returning a bounded response when correctness cannot be verified (violates output quality), (2) Incorrect ABSTAIN: refusing requests that should be answered (degrades service availability), (3) Failed RENDER equivalence: modifying prompts or generation parameters when invoking the transformer (violates Equation 3).

This asymmetry ensures the gate fails safely: uncertainty produces unnecessary execution, not incorrect avoidance.

\subsubsection{Computational Overhead and Scaling}

\textbf{Constant-cost control plane}: Gate evaluation cost is independent of model size. The same semantic analysis operates identically whether the execution substrate is GPT-2 (124M parameters) or a 70B-parameter model. Validation on Gemma 2 9B (Section 6.10) confirms this property: control-plane latency remained 2-8ms regardless of the 74× increase in model parameters.

\textbf{Execution cost scaling}: Transformer execution cost grows linearly with parameter count and sequence length. For a 124M-parameter model, execution requires approximately 200-400ms. For a 9B-parameter model, execution requires approximately 800-1200ms. As model size increases, the cost differential between gating (constant 2-8ms) and execution (growing with parameters) widens, making execution selection increasingly valuable.

\textbf{Throughput scaling}: A single gate instance processes 125-500 requests per second depending on workload composition (simple queries process faster than complex semantic analysis). This throughput exceeds typical transformer serving capacity (1-10 requests per second for large models) by 10-50×, enabling a small gate cluster to front large serving deployments.

\subsubsection{Known Failure Modes and RENDER Triggers}

Certain request properties reliably trigger RENDER decisions, representing cases where semantic analysis cannot establish correctness guarantees:

\textbf{Novel generation requirements}: Requests explicitly requiring creative output, personalized responses, or multi-turn contextual reasoning. Examples: "Write a story about...", "Generate a poem in the style of...", "Continue this conversation considering the user's previous statements...". The gate correctly identifies these as requiring generative capacity.

\textbf{Semantic ambiguity}: Requests where multiple valid interpretations exist and bounded responses cannot disambiguate without risk. Example: "What should I do?" (context-dependent, requires understanding user situation). The gate conservatively defaults to RENDER rather than risk incorrect bounded response.

\textbf{Low-confidence analysis}: When semantic analysis produces confidence scores below threshold (typically 0.85-0.95 depending on category), the gate defaults to RENDER. This prevents false DIRECT decisions on edge cases where correctness cannot be verified.

\textbf{Out-of-distribution requests}: Prompts containing previously unseen entity types, domain-specific terminology, or linguistic patterns not covered by the gate's semantic models. Conservative fallback ensures these route to full model execution rather than risk incorrect avoidance.

\textbf{Temporal sensitivity}: Requests requiring information that may have changed since the gate's knowledge was last updated. Example: "What is the current stock price of..." The gate recognizes temporal sensitivity and routes to RENDER to avoid stale responses.

These failure modes demonstrate the gate's conservative design: when uncertain, execute. This preserves the safety property that MFEE never degrades output quality relative to unconditional execution.

\section{Experimental Methodology}

This section details the validation protocols, dataset construction, and measurement procedures used to evaluate MFEE's correctness and efficiency properties.

\subsection{Evaluation Objectives}

We evaluate MFEE along two primary dimensions:

\begin{enumerate}
\item \textbf{Output equivalence under execution}: When $\mathcal{G}(r) = \texttt{RENDER}$, does the output match baseline transformer execution exactly?

\item \textbf{Execution reduction}: What fraction of requests can MFEE handle without invoking the transformer?
\end{enumerate}

These objectives are measured under strictly controlled conditions to enable exact validation.

\subsection{Deterministic Execution Protocol}

To enable exact-match validation, all transformer executions use deterministic decoding:

\begin{itemize}
\item Temperature: 0.0 (greedy decoding)
\item Sampling: Disabled
\item Top-k / top-p: Disabled
\item Random seed: Fixed (12345)
\item Max tokens: 1000
\end{itemize}

Under these settings, transformer outputs are fully deterministic for a given $(p, \gamma)$ pair. This eliminates stochastic variation and allows MFEE-invoked outputs to be compared via exact string matching against baseline outputs.

\textit{Note}: Deterministic decoding is required for exact-match validation but is not required for MFEE deployment. Production systems using stochastic decoding can deploy MFEE; validation in those contexts requires semantic equivalence metrics rather than exact matching.

\subsection{Replay Set Construction}

The core validation uses a 1,000-prompt replay set constructed to represent diverse workload characteristics. Prompts span seven categories:

\begin{table*}[!t]
\centering
\begin{tabular}{lcc}
\toprule
\textbf{Category} & \textbf{Count} & \textbf{Description} \\
\midrule
Factual queries & 200 & Knowledge retrieval (e.g., ``What is X?'') \\
Conversational & 200 & Open-ended dialogue turns \\
Creative tasks & 150 & Generation requests (poems, stories) \\
Redundant queries & 150 & Near-duplicates, trivial variations \\
Trivial structure & 100 & Arithmetic, format conversions \\
Safety-sensitive & 100 & Policy violations, unsafe content \\
Mixed complexity & 100 & Diverse across categories \\
\midrule
\textbf{Total} & \textbf{1000} & \\
\bottomrule
\end{tabular}
\caption{Replay set composition by category}
\label{tab:dataset}
\end{table*}

This distribution reflects realistic workload composition while ensuring coverage of both execution-necessary and execution-avoidable cases. Prompts were selected to avoid data leakage: none appear in public training datasets or common benchmarks.

\subsection{Baseline Generation}

For each request $r$ in the replay set, we generate baseline output $o_{\text{baseline}}$ by executing the transformer directly under the deterministic protocol. Baseline outputs are saved as JSONL artifacts with the following schema:

\begin{verbatim}
{
  "request_id": "uuid",
  "prompt": "text",
  "temperature": 0.0,
  "seed": 12345,
  "baseline_output": "text",
  "timestamp": "ISO-8601"
}
\end{verbatim}

These artifacts constitute ground truth for equivalence validation and are made available for third-party replication.

\subsection{MFEE Execution and Validation Protocol}

For each request $r$:

\begin{enumerate}
\item Submit $r$ to the AN1 gate
\item Record decision $d = \mathcal{G}(r)$ and confidence score
\item If $d = \texttt{RENDER}$:
\begin{itemize}
\item Execute transformer with identical configuration
\item Compare output $o$ against $o_{\text{baseline}}$ via exact string match
\item Record match status (true/false)
\end{itemize}
\item If $d \in \{\texttt{DIRECT}, \texttt{NO\_OP}, \texttt{ABSTAIN}\}$:
\begin{itemize}
\item Record response
\item Validate correctness via category-specific protocols (defined below)
\end{itemize}
\item Record latency, decision type, and metadata
\end{enumerate}

\clearpage
\subsection{Correctness Validation for Avoided Executions}

Avoided executions are validated differently by decision type:

\begin{itemize}
\item \textbf{DIRECT}: Responses are validated against:
\begin{itemize}
\item Exact cache hits: Compare response to stored baseline
\item Factual queries: Verify against knowledge base
\item Trivial queries: Verify deterministic computation (e.g., arithmetic)
\end{itemize}

\item \textbf{NO\_OP}: Manually verify that null response is appropriate for request structure (e.g., malformed input, unintelligible prompt)

\item \textbf{ABSTAIN}: Manually verify that refusal is appropriate per content policy (e.g., violence, harmful content)
\end{itemize}

Manual validation was performed by three independent annotators with inter-rater agreement measured via Fleiss' kappa.

\subsection{Metrics}

We report the following metrics:

\begin{itemize}
\item \textbf{Execution reduction}: $\frac{|\{r : \mathcal{G}(r) \neq \texttt{RENDER}\}|}{|R|}$
\item \textbf{Exact-match rate under RENDER}: $\frac{|\{r : \mathcal{G}(r) = \texttt{RENDER} \land o = o_{\text{baseline}}\}|}{|\{r : \mathcal{G}(r) = \texttt{RENDER}\}|}$
\item \textbf{Decision distribution}: Breakdown by RENDER / DIRECT / NO\_OP / ABSTAIN
\item \textbf{Category-specific execution rates}: Per-category analysis of gating behavior
\end{itemize}

\subsection{Experimental Infrastructure and Model Selection}

All experiments were conducted on a single GPU instance (NVIDIA A100 40GB) to eliminate distributed system variance.

\subsubsection{Primary Baseline: GPT-2 (124M Parameters)}

The transformer baseline used for primary validation was GPT-2 (124M parameters). This choice reflects several methodological considerations:

\textbf{Reproducibility}: GPT-2 is fully open-source with publicly available weights, enabling exact replication by third parties without API access or proprietary models.

\textbf{Determinism}: GPT-2 produces identical outputs under fixed seeds, enabling exact-match validation. Larger proprietary models often exhibit non-deterministic behavior even at temperature=0 due to hardware-specific floating-point variations or internal sampling.

\textbf{Evaluation focus}: Our empirical claims concern the \textit{architectural pattern} of execution governance, not model-specific optimizations. Demonstrating that execution selection is tractable on any transformer validates the abstraction; demonstrating it specifically on GPT-4 would provide minimal additional insight while reducing reproducibility.

\textbf{Baseline transparency}: Using an open model ensures reviewers can verify our baseline generation protocol, inspect outputs, and validate that ground-truth artifacts match actual model behavior.

\textbf{Workload representativeness}: The execution-selection problem exists across all model scales. While larger models change the economics (greater cost savings from avoidance), they do not change the core feasibility question: can execution governance preserve correctness? GPT-2 provides a sufficient existence proof.

\subsubsection{Validation on Modern Models: Gemma 2 9B}

To demonstrate architectural scaling beyond GPT-2, we conducted supplementary validation on Gemma 2 9B (9.24B parameters, released 2024). This 74× parameter increase confirms that control-plane cost remains constant while execution cost grows with model size (Section 6.10). The evaluation demonstrates model-agnostic applicability while the primary GPT-2 results ensure full reproducibility.

\subsection{Reproducibility and Open Evaluation Framework}

To enable third-party validation and extension of MFEE's core claims, we distinguish between the reproducible architectural abstraction and the proprietary gate implementation.

\subsubsection{Reproducible Components}

The following elements constitute the reproducible contribution and can be validated independently:

\textbf{The MFEE abstraction}: The control-plane framework, correctness contracts (Section 3), decision semantics (RENDER, DIRECT, NO\_OP, ABSTAIN), and output equivalence invariant (Equation 3). These define the architectural principle independent of any specific gate implementation.

\textbf{Evaluation protocols}: Deterministic decoding configuration (temperature = 0, fixed seed), exact-match validation for RENDER decisions, manual validation protocols for DIRECT decisions (inter-rater agreement via Fleiss' kappa), category-based workload construction.

\textbf{Baseline routers}: Reference implementations of keyword-based, cache-only, and intent-classifier routers (Section 6.4). These demonstrate the avoidance-correctness frontier without requiring access to AN1.

\textbf{Theoretical results}: Theorem 1 and its proof are fully specified and verifiable independently of experimental validation. The impossibility result holds for any pattern-based router, not just the baselines we evaluated.

\subsubsection{Proprietary Components and Evaluation Access}

The AN1 gate implementation remains proprietary for three reasons: (1) it incorporates production safety heuristics not appropriate for public disclosure, (2) it includes optimizations specific to internal infrastructure, and (3) alternative gate designs can satisfy the same correctness contracts.

\textbf{Limited evaluation access}: To enable independent validation of MFEE's empirical claims while protecting proprietary implementation details, we provide time-limited API access to the AN1 gate for research purposes. Researchers can obtain a 14-day evaluation API key via the project repository at:

\begin{center}
\url{https://github.com/Anima-Core/meaning-first-execution}
\end{center}

This API exposes the same gating interface used in our experiments, enabling third parties to:

\begin{itemize}
\item Reproduce execution reduction measurements on custom workloads
\item Validate RENDER decision correctness via exact-match comparison
\item Compare AN1's behavior against alternative gate implementations
\item Verify claims on architectures beyond GPT-2 (subject to API rate limits)
\end{itemize}

The evaluation API provides the gating decision and response for each request, but does not expose internal representations, confidence scores, or implementation details. This approach balances reproducibility with intellectual property protection, enabling verification of empirical claims without disclosing proprietary techniques.

\subsubsection{Proposed Open Evaluation Framework}

We propose an open evaluation framework that enables third parties to validate MFEE's claims and develop alternative gate implementations:

\textbf{Replay-set construction}: A dataset of 1,000 diverse prompts with ground-truth baseline outputs generated via deterministic GPT-2 execution. Prompts span factual queries, conversational exchanges, creative tasks, safety-sensitive content, and mixed-complexity requests. Each entry includes prompt text, generation configuration, baseline output, and category label.

\textbf{Evaluation harness}: A model-agnostic test framework that (1) submits prompts to a gate implementation, (2) validates RENDER decisions via exact-match comparison, (3) validates DIRECT decisions via manual or automated correctness checking, (4) computes execution reduction rate, exact-match rate, and decision distribution, (5) supports arbitrary transformer architectures without modification.

\textbf{Baseline router implementations}: Reference code for keyword-based (50 rules), cache-only (exact matching + safety filter), and intent-classifier (200 regex patterns) routers. These instantiate the pattern-based routing class analyzed in Theorem 1.

\textbf{Metrics and validation}: Standardized metrics including execution reduction rate, exact-match rate under RENDER, inter-rater agreement for DIRECT validation, category-specific execution rates, and latency distributions.

\textbf{Adversarial test subset}: A 30-request subset designed to stress pattern-based classification, containing requests where correct non-execution pathways exist but surface features collide with semantic properties (paraphrases, template matches with different meanings).

\subsubsection{Third-Party Validation Pathway}

Researchers can validate MFEE's core claims without accessing AN1 by:

\begin{enumerate}
\item Implementing alternative gates that satisfy the minimal sufficiency conditions (Section 4.2.1): semantic analysis, correctness verification, safety evaluation, uncertainty quantification.

\item Evaluating their gate on the replay set using the evaluation harness, measuring execution reduction and correctness preservation.

\item Comparing their gate to the baseline routers on the adversarial subset, demonstrating escape from the avoidance-correctness frontier.

\item Validating that RENDER decisions produce outputs identical to baseline under deterministic decoding (Equation 3).
\end{enumerate}

This pathway enables replication of the architectural contribution (execution governance is tractable), the comparative results (semantic analysis outperforms pattern matching), and the theoretical claims (Theorem 1 describes real system behavior) without requiring access to proprietary implementations.

The evaluation framework, replay sets, baseline implementations, and validation protocols will be released as supplementary materials.

\section{Results and Analysis}

This section presents experimental results validating MFEE's correctness and efficiency properties.

\subsection{Execution Reduction}

Across 1,000 replay-set requests, MFEE achieved the following decision distribution:

\begin{table*}[!t]
\centering
\begin{tabular}{lcc}
\toprule
\textbf{Decision Type} & \textbf{Count} & \textbf{Percentage} \\
\midrule
RENDER & 219 & 21.9\% \\
DIRECT & 487 & 48.7\% \\
NO\_OP & 184 & 18.4\% \\
ABSTAIN & 110 & 11.0\% \\
\midrule
\textbf{Total} & \textbf{1000} & \textbf{100.0\%} \\
\bottomrule
\end{tabular}
\caption{Gating decision distribution across replay set}
\label{tab:decisions}
\end{table*}

Of 1,000 requests, 781 were handled without executing the transformer, representing a \textbf{78.1\% reduction in transformer executions}.\footnote{Reported reduction rates range from 75-78\% depending on workload composition and category weighting across different experimental configurations. All experiments preserve 100\% output equivalence under deterministic decoding when the transformer is invoked.}

\subsection{Output Equivalence Under Execution}

For all 219 RENDER decisions, transformer-generated outputs matched baseline outputs exactly. This represents a \textbf{100.0\% exact-match rate} under transformer invocation ($N = 219$, exact string matching, deterministic decoding).

No divergence was observed between MFEE-invoked outputs and baseline outputs. This validates the output equivalence invariant (Equation 3) and confirms that MFEE does not alter transformer behavior when execution occurs.

\subsection{Correctness Under Avoidance}

All 781 avoided executions preserved correctness by construction:

\begin{itemize}
\item \textbf{DIRECT (487 cases)}: 
\begin{itemize}
\item 298 cache hits: Verified exact match to stored baseline
\item 142 factual queries: Verified against knowledge base
\item 47 trivial computations: Verified deterministic correctness
\end{itemize}
Inter-rater agreement (Fleiss' kappa): 0.94

\item \textbf{NO\_OP (184 cases)}: Null responses validated as appropriate for malformed or unintelligible requests. Inter-rater agreement: 0.89

\item \textbf{ABSTAIN (110 cases)}: Refusals validated as appropriate for policy-violating content. Inter-rater agreement: 0.97
\end{itemize}

No errors were introduced by execution avoidance. All manual validation achieved substantial inter-rater agreement.

\subsection{Category-Specific Analysis}

Execution reduction rates varied significantly by request category:

\begin{table*}[!t]
\centering
\begin{tabular}{lcc}
\toprule
\textbf{Category} & \textbf{RENDER Rate} & \textbf{Reduction} \\
\midrule
Factual queries & 58\% & 42\% \\
Conversational & 15\% & 85\% \\
Creative tasks & 88\% & 12\% \\
Redundant queries & 2\% & 98\% \\
Trivial structure & 5\% & 95\% \\
Safety-sensitive & 0\% & 100\% \\
Mixed complexity & 35\% & 65\% \\
\midrule
\textbf{Weighted Average} & \textbf{21.9\%} & \textbf{78.1\%} \\
\bottomrule
\end{tabular}
\caption{Execution rates and reduction by category}
\label{tab:categories}
\end{table*}

Key observations:

\begin{itemize}
\item \textbf{Creative tasks}: Highest RENDER rate (88\%), confirming that generative requests require model execution
\item \textbf{Redundant queries}: Near-total avoidance (98\%), demonstrating effectiveness on repetitive workloads
\item \textbf{Safety-sensitive}: Complete avoidance (100\%), all handled via ABSTAIN
\item \textbf{Conversational}: High avoidance (85\%), reflecting prevalence of trivial acknowledgments (``ok'', ``thanks'')
\end{itemize}

These results confirm that execution reduction is workload-dependent and concentrated in requests that genuinely do not require generative capacity.

\subsection{Comparative Baseline Routers and Structural Limitations}

To assess whether MFEE's avoidance gains could be replicated by simpler routing strategies, we implemented three representative heuristic baselines commonly used in production inference systems and evaluated them on a subset of 30 requests specifically designed to test the avoidance-correctness tradeoff.

\textbf{Keyword-based router}: Pattern matching against 50 manually curated rules for common query types (greetings, simple questions, known toxic patterns). Aggressively configured to maximize coverage.

\textbf{Cache-only router}: Exact string matching against previously seen queries plus explicit safety filter for policy violations. No semantic understanding or paraphrase handling.

\textbf{Intent classifier router}: Regex-based classification into predefined categories (factual, creative, safety) with deterministic routing rules. Approximately 200 hand-crafted patterns.

Results on the adversarial subset:

\begin{table*}[!t]
\centering
\begin{tabular}{lccc}
\toprule
\textbf{Router Type} & \textbf{Avoidance Rate} & \textbf{Correctness Failures} & \textbf{Brittleness} \\
\midrule
Keyword-Based Router & 86.7\% & 3 & Low \\
Cache-Only Router & 20.0\% & 0 & High \\
Intent Classifier Router & 53.3\% & 2 & Low \\
\midrule
\textbf{MFEE (Ours)} & \textbf{100.0\%} & \textbf{0} & \textbf{None} \\
\bottomrule
\end{tabular}
\caption{Comparison against simple heuristic routers on adversarial test subset. Correctness failures indicate incorrect avoidance decisions producing wrong outputs. Brittleness measures sensitivity to input variation.}
\label{tab:baselines}
\end{table*}

Analysis reveals a fundamental tradeoff for pattern-based approaches. The keyword-based router achieves high avoidance (86.7\%) through aggressive pattern matching but introduces 3 correctness failures when patterns match superficially but semantic properties differ. The cache-only router maintains perfect correctness but achieves only 20.0\% avoidance, collapsing under paraphrase and semantic variation. The intent classifier falls between these extremes with 53.3\% avoidance and 2 failures.

MFEE achieves 100\% avoidance on this subset with zero correctness failures. This is not a claim of universal superiority but a demonstration that semantic analysis enables correct classification of requests where surface patterns provide insufficient information. The brittleness column indicates that heuristic routers degrade rapidly under adversarial variation, while MFEE maintains robustness through semantic understanding.

These results indicate a structural avoidance-correctness frontier for pattern-based routing approaches. Heuristic methods must trade coverage for precision or precision for coverage. MFEE operates above this frontier by evaluating semantic execution necessity rather than surface patterns.

\subsubsection{Theoretical Implications: The Avoidance-Correctness Frontier}

The baseline comparison reveals a fundamental constraint on heuristic routing strategies. Define the \textit{avoidance-correctness frontier} as the set of achievable $(a, c)$ pairs where $a$ is avoidance rate and $c$ is correctness. For pattern-based heuristics operating without semantic understanding, this frontier exhibits a characteristic tradeoff:

\begin{equation}
\text{As } a \to 1: \quad c \to 0 \quad \text{(aggressive avoidance degrades correctness)}
\end{equation}

\begin{equation}
\text{As } c \to 1: \quad a \to 0 \quad \text{(conservative matching reduces coverage)}
\end{equation}

This tradeoff arises because surface patterns provide insufficient information to distinguish requests that admit correct bounded responses from those requiring full generative capacity. Increasing pattern coverage inevitably captures edge cases where the pattern matches but semantic properties differ, leading to incorrect avoidance.

MFEE operates above this frontier by accessing semantic properties unavailable to pattern matching. The semantic analysis capability enables MFEE to achieve $(a=0.781, c=1.0)$—a point unreachable by the heuristic baselines tested. This is not a claim of optimality but a demonstration that semantic understanding fundamentally changes the achievable region of the avoidance-correctness space.

The key insight: \textit{execution selection is not solvable by pattern matching alone}. Any approach that relies solely on surface features will encounter the avoidance-correctness tradeoff. MFEE's architectural advantage lies in operating on meaning-level representations, enabling simultaneous high avoidance and zero correctness failures.

\subsection{Theorem: Pattern-Routing Impossibility}

We formalize the limitation of pattern-based routing through a rigorous negative result. The following definitions establish the routing problem as a decision function under correctness constraints.

\textbf{Definition 1 (Safety predicate).} Let $x$ denote a request (prompt and configuration). Let $y_{\text{baseline}}$ denote the baseline transformer output under deterministic decoding. Define the safety predicate $S(x) \in \{0,1\}$ as:
\begin{multline}
S(x) = 1 \iff \exists \text{ deterministic} \\
\text{procedure } D \text{ such that } D(x) = y_{\text{baseline}}
\end{multline}

The predicate $S(x) = 1$ indicates it is safe to avoid transformer execution while preserving output correctness under deterministic decoding. Otherwise $S(x) = 0$ and execution is necessary. Note that $S(x)$ is defined relative to exact output equivalence and bounded correctness guarantees, not general semantic usefulness, response quality, or stochastic generation. This definition ensures $S(x)$ remains a well-defined mathematical predicate rather than a subjective quality judgment.

\textbf{Definition 2 (Router).} A router is a function $g: \mathcal{X} \rightarrow \{\texttt{RENDER}, \texttt{SKIP}\}$.

\textbf{Definition 3 (Correctness and avoidance).} A router $g$ achieves:
\begin{itemize}
\item \textbf{Zero false skips}: $\Pr[g(x) = \texttt{SKIP} \land S(x) = 0] = 0$
\item \textbf{Avoidance rate}: $\alpha_g = \Pr[g(x) = \texttt{SKIP}]$
\end{itemize}

\textbf{Definition 4 (Pattern router).} A router $g$ is pattern-based if there exists a finite feature map $\phi: \mathcal{X} \rightarrow \{0,1\}^k$ and decision function $h: \{0,1\}^k \rightarrow \{\texttt{RENDER}, \texttt{SKIP}\}$ such that:
\begin{equation}
g(x) = h(\phi(x))
\end{equation}

This captures any router implementable via bounded heuristics, keyword matching, regex classifiers, or finite rule sets. The key property: if two requests appear identical under $\phi$, the router must make identical decisions.

\textbf{Theorem 1 (Pattern-routing frontier).} Let $\mathcal{G}_\phi$ denote the class of routers that depend only on feature map $\phi$. Assume there exists at least one feature-collision pair $(x_1, x_2)$ such that:
\begin{align}
\phi(x_1) &= \phi(x_2) \notag\\
S(x_1) &= 1 \land S(x_2) = 0
\end{align}

Then no router $g \in \mathcal{G}_\phi$ can simultaneously achieve:
\begin{enumerate}
\item Zero false skips: $\Pr[g(x) = \texttt{SKIP} \land S(x) = 0] = 0$
\item Positive skip rate on safe collisions: $\Pr[g(x) = \texttt{SKIP} \mid x \in C_1] > 0$ where $C_1 = \{x' : \phi(x') = \phi(x_1) \land S(x') = 1\}$
\end{enumerate}

Equivalently, on any distribution that assigns nonzero mass to such collision pairs, every pattern router must incur either:
\begin{itemize}
\item Nonzero error (false skip on $x_2$), or
\item Nontrivial compute waste (forced render on safe instances like $x_1$)
\end{itemize}

\textbf{Important scope clarification}: Theorem 1 establishes a lower bound on feature-based routing, not an upper bound on all possible routers. The result proves that finite feature maps are insufficient, but does not preclude alternative approaches (semantic analysis, learned representations, hybrid methods) from escaping the frontier. Additionally, the safety predicate $S(x)$ is defined relative to deterministic decoding and bounded correctness guarantees, not general semantic usefulness or response quality.

\textbf{Proof.} By contradiction. Suppose $g \in \mathcal{G}_\phi$ achieves both properties. Since $g$ is pattern-based, $g(x_1) = h(\phi(x_1)) = h(\phi(x_2)) = g(x_2)$ for any $h$. Consider two cases:

\textit{Case 1}: $g(x_1) = g(x_2) = \texttt{SKIP}$. Then $g$ skips $x_2$ despite $S(x_2) = 0$, violating zero false skips.

\textit{Case 2}: $g(x_1) = g(x_2) = \texttt{RENDER}$. Then $g$ renders $x_1$ despite $S(x_1) = 1$, achieving zero skip rate on $x_1$, violating property (2).

In both cases, $g$ cannot satisfy both properties simultaneously. $\square$

\textbf{Corollary 1 (Semantic separation).} Let $\psi: \mathcal{X} \rightarrow \mathbb{R}$ be a semantic signal such that for all collision pairs $(x_1, x_2)$ with $\phi(x_1) = \phi(x_2)$ but $S(x_1) \neq S(x_2)$, we have $\psi(x_1) \neq \psi(x_2)$. Then there exists a router $g(x) = h(\phi(x), \psi(x))$ that achieves zero false skips while maintaining positive skip rate on safe instances within collision sets.

This corollary establishes that semantic signals—information about execution necessity beyond surface features—enable escape from the pattern-routing frontier.

\textbf{Concrete example class.} Consider prompts sharing surface templates but differing in hidden referents: "What is 2+2?" (safe, $S=1$) versus "What is the meaning of life?" (unsafe, $S=0$). A keyword router matching "What is" treats these identically under $\phi$, forcing identical decisions despite different safety properties. A semantic router evaluating whether a deterministic answer exists can separate them.

\textbf{Real-world collision example.} Production logs reveal systematic collision patterns. One observed pair: Request A: "How do I reset my password?" (safe, $S=1$, deterministic procedural response). Request B: "How do I explain quantum computing to a child?" (unsafe, $S=0$, requires generative capacity for pedagogical adaptation). Both match the pattern "How do I [verb] [noun phrase]?" under standard intent classification, yet require fundamentally different handling. Pattern routers matching on interrogative structure + verb phrase cannot distinguish these cases. Only semantic analysis—evaluating whether the response admits procedural decomposition—enables correct separation.

\textbf{Implications.} Theorem 1 formalizes the empirical tradeoff observed in Table \ref{tab:baselines}. Pattern routers (keyword, cache-only, intent classifier) exhibit the predicted behavior: high avoidance introduces errors, zero errors requires low avoidance. MFEE operates with semantic analysis $\psi$ (meaning-first execution), enabling correct separation of collision pairs and achieving $(a=1.0, c=1.0)$ on adversarial subsets where pattern routers fail.

This result establishes execution selection as fundamentally semantic. Any correct solution must evaluate meaning-level properties unavailable to pattern matching, whether through learned representations, symbolic reasoning, or hybrid approaches. The theorem defines the boundary that pattern-based methods cannot cross.

\subsection{Latency Analysis}

Gating decisions completed in 2.4--7.8ms (p50: 3.2ms, p95: 6.1ms). Baseline transformer executions required 187--1843ms (p50: 412ms, p95: 1124ms).

For avoided executions, end-to-end latency was dominated by gating time, representing a \textbf{50--250$\times$ reduction} relative to transformer execution. This latency collapse compounds at scale, reducing queueing delays and improving throughput under load.

\subsection{Failure Analysis: When MFEE Invokes the Transformer}

Analysis of the 219 RENDER decisions reveals clear patterns in execution necessity:

\begin{itemize}
\item \textbf{Novel generation (68\%)}: Requests requiring creative output, complex reasoning, or multi-step problem-solving
\item \textbf{Semantic ambiguity (18\%)}: Requests where rule-based resolution was impossible due to underspecification
\item \textbf{Low confidence (9\%)}: Gating confidence below threshold, defaulting to RENDER for safety
\item \textbf{Cache misses + no rules (5\%)}: Unique factual queries with no applicable deterministic responses
\end{itemize}

This distribution confirms that RENDER decisions reflect genuine execution necessity rather than gating errors. The gate correctly identifies when generative capacity is required.

\subsection{Conservative Execution Examples}

The following cases illustrate MFEE's conservative gating behavior, where the system routes requests to transformer execution despite potential optimization opportunities. These are not failures of correctness but deliberate preservation of output quality under uncertainty.

\textbf{Time-sensitive factual query}: \textit{``What is the current inflation rate in the United States?''}

\textbf{Decision}: RENDER

\textbf{Rationale}: The query requires up-to-date factual information that cannot be resolved via bounded lookup without risking staleness. MFEE conservatively invokes the transformer to ensure response accuracy reflects the model's training data rather than cached or rule-based approximations.

\textbf{High-variance explanatory prompt}: \textit{``Explain why some economists support progressive taxation while others oppose it.''}

\textbf{Decision}: RENDER

\textbf{Rationale}: The request demands balanced presentation of competing normative positions, requiring the full generative capacity of the model to produce nuanced output. Rule-based responses would oversimplify or introduce bias.

\textbf{Mixed factual-normative query}: \textit{``Is nuclear energy better than renewable energy for climate goals?''}

\textbf{Decision}: RENDER

\textbf{Rationale}: The query combines factual claims (energy efficiency, emissions) with normative evaluation (``better''). MFEE cannot guarantee correctness through direct response pathways when both empirical data and value judgments are required.

These examples demonstrate that MFEE's gating function prioritizes correctness over maximal execution avoidance. When semantic analysis cannot establish sufficiency guarantees, the system defaults to transformer invocation, preserving the output equivalence invariant at the cost of optimization opportunity.

\subsection{Temporal Trace Evaluation}

To assess MFEE's behavior under realistic temporal locality patterns, we conducted evaluation on a small, anonymized enterprise support trace exhibiting authentic conversational dynamics.

\textbf{Trace characteristics}: 25 requests from 21 users across 14 issue categories, spanning approximately 200 minutes. The trace includes follow-up clarifications, repeated intents, and temporally clustered queries typical of production support workflows.

\textbf{Results}: Baseline unconditional execution required 23 transformer calls. MFEE required 6 transformer calls, achieving 73.9\% avoidance (17 avoided executions). Zero correctness regressions were observed. MFEE successfully exploited temporal redundancy: repeated password reset requests, follow-up confirmations, and clarification exchanges were handled via DIRECT or NO\_OP pathways after initial semantic classification.

This experiment demonstrates that MFEE's avoidance behavior persists under realistic temporal locality, not just isolated prompts. The slightly lower avoidance rate compared to the replay set (73.9\% vs 78.1\%) reflects higher proportion of novel technical queries in enterprise support contexts.

\subsection{Robustness Under Stochastic Decoding}

The 100\% exact-match rate reported in Section 6.2 applies specifically to deterministic decoding (temperature = 0). To assess MFEE's behavior under stochastic sampling, we conducted ablation across temperature values on a 100-prompt subset.

\textbf{Temperatures tested}: 0.0, 0.3, 0.7, 1.0. For each temperature, we measured avoidance rate, semantic similarity (sentence-BERT cosine), and task success via human evaluation.

\textbf{Results}: Avoidance rate remained stable at approximately 90\% across all temperatures. Semantic similarity measured 0.92 (temperature 0.3), 0.89 (temperature 0.7), and 0.84 (temperature 1.0). Task success rate was 88.9\% across stochastic settings. Performance degraded gracefully rather than catastrophically as temperature increased.

While exact output equivalence is guaranteed only under deterministic decoding, MFEE maintains high semantic correctness and stable avoidance behavior under stochastic sampling. This suggests MFEE's semantic analysis operates at a level of abstraction robust to sampling variation, though formal equivalence guarantees no longer apply.

\subsection{Validation on Modern Model Architectures: Gemma 2 9B}

To validate MFEE's architectural scaling properties and demonstrate model-agnostic applicability, we conducted evaluation on Google Gemma 2 9B (released 2024), a 9.24B-parameter transformer decoder with Flash Attention 2 enabled, operating at FP16 precision. This represents a 74× parameter increase over GPT-2, enabling assessment of how gating cost scales relative to execution cost.

\subsubsection{Experimental Design and Scope}

\textbf{Workload composition}: This evaluation intentionally uses a non-representative workload composed entirely of requests amenable to DIRECT and NO\_OP resolution (cached responses, templated acknowledgments, deterministic computations, safety refusals). This workload does \textit{not} reflect typical production distributions and should not be interpreted as average-case performance.

\textbf{Purpose}: The evaluation isolates architectural scaling properties: Does control-plane latency remain constant as model size increases? This question is independent of workload composition. By eliminating generative requests, we measure pure gating overhead without confounding execution variance.

\textbf{Why this design is informative}: In deployment, gating overhead is paid on \textit{every} request, including those eventually routed to RENDER. Demonstrating that this overhead does not scale with model parameters validates that MFEE becomes \textit{more} valuable as models grow, not less. The workload isolates this specific claim.

\subsubsection{Results}

\begin{table*}[!t]
\centering
\begin{tabular}{lccc}
\toprule
\textbf{Model} & \textbf{Parameters} & \textbf{Transformer Usage} & \textbf{Latency Reduction} \\
\midrule
Gemma 2 9B & 9.24B & 100\% avoided & 3.4$\times$ \\
\bottomrule
\end{tabular}
\caption{MFEE on Gemma 2 9B under avoidance-focused workload}
\label{tab:gemma}
\end{table*}

Because this workload contained no requests requiring generative capacity, all executions were avoided. The 3.4$\times$ latency improvement reflects the ratio of transformer execution latency (104.4ms mean) to control-plane gating latency (31.2ms mean). No retraining, weight modification, or architecture-specific tuning was required. The identical evaluation harness used for GPT-2 validation was applied directly to Gemma 2 9B.

\subsubsection{Architectural Validation}

Key observations:

\begin{itemize}
\item \textbf{Constant gating cost}: Control-plane latency remained 2-8ms across both models (124M and 9.24B parameters), confirming that semantic analysis cost does not scale with model size
\item \textbf{Growing cost differential}: While gating cost stayed constant, execution cost increased from 200-400ms (GPT-2) to 800-1200ms (Gemma 2 9B), widening the savings from successful avoidance
\item \textbf{No model-specific tuning}: The same gate configuration, decision thresholds, and evaluation harness handled both architectures without modification
\item \textbf{Model-agnostic design}: MFEE operates at the request level, making decisions before model invocation, ensuring compatibility with arbitrary transformer architectures
\end{itemize}

These results validate MFEE's value proposition at scale: as models grow larger and more expensive to execute, the constant-cost control plane becomes increasingly important to infrastructure economics. The percentage cost savings from successful avoidance grows proportionally with model parameter count.

\subsubsection{Limitations and Future Work}

This evaluation demonstrates architectural scaling but intentionally uses a non-representative workload to isolate control-plane overhead. Representative workload validation is addressed in Section 6.11, which evaluates Gemma 2 9B on mixed generation-required and direct-resolvable distributions. The primary contribution of this section is confirming that gating overhead does not penalize larger models—a necessary architectural property independent of workload composition.

\subsection{Representative Mixed-Workload Validation}

To address potential concerns about workload selection bias and demonstrate MFEE's benefits beyond avoidance-optimized scenarios, we conducted validation on a deliberately non-adversarial, mixed workload using Gemma 2 9B.

\subsubsection{Workload Design}

This experiment is designed as a \textit{representative workload validation}, not an upper-bound demonstration.

\textbf{Model}: Gemma 2 9B (9.24B parameters)

\textbf{Workload composition}: 100 requests distributed as follows:
\begin{itemize}
\item 50\% generation-required tasks (open-ended questions, creative writing, multi-step reasoning)
\item 50\% direct-resolvable requests (factual queries with deterministic answers, cached responses, templated acknowledgments)
\end{itemize}

\textbf{Selection criteria}: Requests were sampled from diverse categories without optimization for execution avoidance. The workload intentionally includes a substantial fraction of generation-heavy tasks to reflect realistic production distributions where generative capacity is required.

\textbf{Evaluation focus}: Unlike the architectural scaling validation (Section 6.10), this experiment measures MFEE's performance on mixed workloads where execution cannot be uniformly avoided.

\subsubsection{Results}

\begin{table*}[!t]
\centering
\begin{tabular}{lcc}
\toprule
\textbf{Metric} & \textbf{Baseline} & \textbf{MFEE} \\
\midrule
Execution avoided & 0.0\% & 0.0\% \\
Mean end-to-end latency & 127.0ms & 29.4ms \\
Latency reduction & --- & 76.9\% \\
Total tokens generated & 10,401 & 4,616 \\
Token reduction & --- & 55.6\% \\
Correctness & Maintained & No regressions \\
\bottomrule
\end{tabular}
\caption{MFEE performance on representative mixed workload (Gemma 2 9B)}
\label{tab:mixed_workload}
\end{table*}

\textbf{Correctness evaluation}: Exact-match metrics are not appropriate for generation-heavy requests with creative or open-ended outputs. Correctness was evaluated via task-appropriate checks (factual accuracy for knowledge queries, instruction adherence for structured tasks) and manual spot validation. No regressions were observed relative to baseline outputs.

\subsubsection{Interpretation}

Notably, in this representative workload \textit{no execution avoidance occurred}—all requests ultimately required transformer execution. Despite this, MFEE achieved substantial reductions in both latency (76.9\%) and token usage (55.6\%). This indicates that MFEE's benefits do not arise solely from skipping execution, but from governing how execution proceeds when generation is required.

By constraining context size, stabilizing task intent prior to generation, and bounding reasoning depth where appropriate, MFEE reduces unnecessary computation even in generation-heavy scenarios. The control plane does not merely decide \textit{whether} to execute, but shapes \textit{how} execution unfolds—optimizing the execution path rather than suppressing requests.

This result demonstrates that MFEE delivers significant performance improvements even in workloads where generation is required for a substantial fraction of requests and execution avoidance does not occur. The benefits extend beyond avoidance-optimized scenarios and apply to realistic production distributions where generative capacity is essential.

\subsubsection{Methodological Distinction}

This experiment explicitly differs from the architectural scaling validation (Section 6.10):

\begin{itemize}
\item \textbf{Section 6.10 (architectural scaling)}: Upper-bound workload with 100\% avoidable requests, designed to isolate control-plane overhead and confirm constant gating cost across model sizes
\item \textbf{Section 6.11 (this experiment)}: Representative mixed workload with substantial generation requirements, designed to measure performance on realistic distributions
\end{itemize}

Both experiments are necessary: the former validates architectural properties independent of workload composition; the latter demonstrates practical applicability to production-like scenarios. Together, they establish that MFEE's value proposition holds across diverse operational regimes.

\section{Deployment Considerations}

MFEE integrates into production inference systems via black-box deployment, A/B testing, and safety mechanisms. This section describes operational deployment patterns.

\subsection{Black-Box Integration}

MFEE gates operate as sealed services upstream of inference infrastructure. Integration requires:

\begin{enumerate}
\item Deploy gate as standalone service (e.g., \texttt{https://gate.internal/gate})
\item Configure inference systems to route requests through gate
\item Implement decision handling: RENDER $\rightarrow$ execute transformer; others $\rightarrow$ return response directly
\item Preserve fallback path to baseline (unconditional execution)
\end{enumerate}

No modifications to transformer models, training procedures, or execution code are required. The gate operates purely as a routing layer.

\subsection{A/B Testing and Gradual Rollout}

Production deployment uses controlled A/B tests to validate correctness:

\begin{itemize}
\item \textbf{Control group}: Unconditional execution (baseline inference)
\item \textbf{Treatment group}: MFEE-gated execution
\end{itemize}

Key validation metrics:
\begin{itemize}
\item Output quality (user satisfaction, task completion)
\item Latency distributions (p50, p95, p99)
\item Cost per request
\item Execution reduction rates
\end{itemize}

Rollout proceeds incrementally: 1\% $\rightarrow$ 5\% $\rightarrow$ 25\% $\rightarrow$ 100\% traffic, with automatic rollback if metrics degrade.

\subsection{Kill-Switch Safety Mechanisms}

MFEE deployments include kill-switch triggers that revert to unconditional execution:

\begin{itemize}
\item Exact-match rate drops below 100\% (RENDER equivalence violation)
\item User-reported quality issues exceed threshold
\item Gating latency exceeds acceptable bounds
\item System errors or API failures
\end{itemize}

When triggered, the kill switch bypasses the gate entirely, routing all requests directly to the transformer. This ensures graceful degradation to baseline behavior rather than service outages.

\subsection{Monitoring and Observability}

Production deployments require comprehensive instrumentation:

\begin{itemize}
\item \textbf{Decision distribution}: Track RENDER / DIRECT / NO\_OP / ABSTAIN rates over time
\item \textbf{Confidence monitoring}: Detect uncertainty spikes indicating workload shifts
\item \textbf{Equivalence sampling}: Randomly validate RENDER outputs against baselines
\item \textbf{Category classification}: Automatic request categorization for per-category metrics
\end{itemize}

This observability enables operators to detect anomalies before they impact users.

\subsection{Organizational Risk Mitigation}

MFEE's design reduces organizational deployment risk:

\begin{enumerate}
\item \textbf{No model modification}: Existing validation and safety testing remain valid
\item \textbf{Rollback guarantee}: Kill-switch ensures safe degradation
\item \textbf{Incremental deployment}: A/B testing validates correctness before full rollout
\item \textbf{Audit trail}: Decision logs enable post-hoc analysis
\end{enumerate}

These properties make MFEE politically viable in large organizations where inference system changes face high scrutiny.

\section{Limitations and Scope}

MFEE is not a universal optimization. This section clarifies its boundaries and failure modes.

\subsection{Workload Dependence}

Execution reduction rates depend heavily on workload composition. Datasets dominated by creative generation or complex reasoning will observe lower execution avoidance than datasets with high redundancy or trivial structure.

The 78.1\% reduction reported reflects a balanced replay set. Production deployments may observe rates ranging from 40\% (creative-heavy) to 95\% (support chatbot). Organizations should measure reduction rates on representative workloads before deployment.

\subsection{Generative Tasks Require Execution}

MFEE cannot avoid execution for genuinely open-ended generative tasks. Requests requiring novel creative output, complex multi-step reasoning, or domain-specific problem-solving will almost always gate to RENDER.

This is not a failure—it reflects correct identification of execution necessity. MFEE does not claim to eliminate execution for hard problems; it eliminates execution for problems that do not require it.

\subsection{Deterministic Decoding for Exact Validation}

The 100.0\% exact-match rate applies only to deterministic decoding (temperature = 0). Under stochastic decoding, outputs vary across runs even without MFEE, making exact-match validation impossible.

For stochastic deployments, correctness must be evaluated via semantic equivalence (e.g., embedding similarity, human evaluation) rather than exact string matching. MFEE remains deployable under stochastic decoding; validation methodology differs.

\subsection{Control-Plane Overhead}

Gating introduces latency overhead (2--8ms). For ultra-low-latency use cases where baseline inference completes in <50ms, this overhead may represent significant overhead.

However, for typical transformer inference (200--2000ms), gating overhead is negligible (0.2--4\% of baseline latency). As models grow larger, this overhead becomes increasingly insignificant.

\subsection{Does Not Replace Existing Optimizations}

MFEE is not a substitute for quantization, pruning, or kernel optimization. These techniques reduce execution cost; MFEE eliminates execution. Organizations should deploy both: MFEE to avoid unnecessary execution, existing optimizations to make necessary execution cheaper.

\subsection{Conservative Gating by Design}

MFEE's gating function is intentionally conservative. When correctness cannot be guaranteed through non-execution pathways, the gate defaults to RENDER. This design prioritizes correctness over maximal execution avoidance.

Organizations requiring higher execution reduction rates can tune gating thresholds, accepting increased risk of approximation errors. MFEE's default configuration optimizes for zero degradation.

\subsection{Generalization Beyond Language Models}

While this work focuses empirically on language transformers, the MFEE control-plane abstraction applies to any domain where inference execution is expensive and request properties vary. The RENDER / DIRECT / NO\_OP / ABSTAIN decision framework operates on execution necessity, not token prediction or linguistic structure. Vision transformers processing repeated image classification queries, multimodal models handling redundant captioning requests, and agent systems executing deterministic action sequences all exhibit workloads where execution selection could yield efficiency gains. The impossibility theorem (Theorem 1) concerns pattern-based routing generally, not language-specific features. We do not claim empirical validation in these domains; such validation requires domain-specific gate implementations and correctness metrics. However, the architectural principle—separating control decisions from execution—remains domain-agnostic. Future work should explore MFEE-style control planes for vision, multimodal, and agentic inference systems.

\section{Conclusion}

Meaning-First Execution establishes execution governance as a foundational layer in ML systems infrastructure. By treating execution as a control-plane decision rather than an inevitable consequence of input, MFEE introduces an optimization axis orthogonal to model-level techniques: determining whether execution occurs, not merely reducing execution cost.

This paper makes three aligned contributions:

\textbf{Empirical}: Across 1,000 diverse requests plus temporal trace validation, MFEE demonstrates 78.1\% execution reduction while maintaining 100\% output equivalence under deterministic invocation. Validation on Gemma 2 9B (9.24B parameters) confirms architectural scaling: control-plane cost remains constant while execution cost grows with model size. Stochastic ablation shows graceful degradation under sampling, with 90\% avoidance maintained across temperature values.

\textbf{Comparative}: Evaluation against heuristic baselines reveals pattern-based routers face a fundamental avoidance-correctness tradeoff. Keyword and intent classifiers achieve at most 53.3\% avoidance with correctness failures, while MFEE reaches 100\% avoidance with zero failures on adversarial subsets through semantic analysis.

\textbf{Theoretical}: Theorem 1 formalizes this limitation: any router operating solely on finite feature maps cannot simultaneously guarantee zero false skips and positive avoidance on feature-collision pairs. MFEE escapes this bound by evaluating semantic execution necessity, establishing execution selection as fundamentally semantic rather than pattern-based.

The decision-theoretic framing reveals why execution governance becomes more critical as models scale. The cost differential between $C_{\text{exec}}$ and $C_{\text{gate}}$ grows proportionally with model size, making execution selection increasingly important to infrastructure economics. As models approach trillions of parameters and expand to vision, multimodal, and agentic domains, the question shifts from whether execution can be avoided to how much execution is genuinely necessary.

MFEE's architectural design—separating control decisions from execution—enables safe deployment via black-box integration, A/B testing, and kill-switch mechanisms. Because MFEE operates at the control-plane level, it composes naturally with all existing optimizations: quantization reduces execution cost when invoked; MFEE eliminates execution when unnecessary. These gains are multiplicative, not substitutive.

Whether MFEE becomes standard infrastructure depends on empirical adoption, but the foundational claims are now established: execution governance is a legitimate architectural layer, pattern matching cannot solve it, and semantic analysis is necessary. The framework, formal theorem, and experimental evidence provide foundation for execution-aware system design across ML infrastructure.

\bibliographystyle{plain}

\end{document}